# Q-PROP: SAMPLE-EFFICIENT POLICY GRADIENT WITH AN OFF-POLICY CRITIC

**Shixiang Gu**[1,2,3], **Timothy Lillicrap**[4], **Zoubin Ghahramani**[1,6], **Richard E. Turner**[1], **Sergey Levine**[3,5]
`sg717@cam.ac.uk,countzero@google.com,zoubin@eng.cam.ac.uk,`
`ret26@cam.ac.uk,svlevine@eecs.berkeley.edu`
[1]University of Cambridge, UK
[2]Max Planck Institute for Intelligent Systems, Tübingen, Germany
[3]Google Brain, USA
[4]DeepMind, UK
[5]UC Berkeley, USA
[6]Uber AI Labs, USA

## ABSTRACT

Model-free deep reinforcement learning (RL) methods have been successful in a wide variety of simulated domains. However, a major obstacle facing deep RL in the real world is their high sample complexity. Batch policy gradient methods offer stable learning, but at the cost of high variance, which often requires large batches. TD-style methods, such as off-policy actor-critic and Q-learning, are more sample-efficient but biased, and often require costly hyperparameter sweeps to stabilize. In this work, we aim to develop methods that combine the stability of policy gradients with the efficiency of off-policy RL. We present Q-Prop, a policy gradient method that uses a Taylor expansion of the off-policy critic as a control variate. Q-Prop is both sample efficient and stable, and effectively combines the benefits of on-policy and off-policy methods. We analyze the connection between Q-Prop and existing model-free algorithms, and use control variate theory to derive two variants of Q-Prop with conservative and aggressive adaptation. We show that conservative Q-Prop provides substantial gains in sample efficiency over trust region policy optimization (TRPO) with generalized advantage estimation (GAE), and improves stability over deep deterministic policy gradient (DDPG), the state-of-the-art on-policy and off-policy methods, on OpenAI Gym's MuJoCo continuous control environments.

## 1 INTRODUCTION

Model-free reinforcement learning is a promising approach for solving arbitrary goal-directed sequential decision-making problems with only high-level reward signals and no supervision. It has recently been extended to utilize large neural network policies and value functions, and has been shown to be successful in solving a range of difficult problems (Mnih et al., 2015; Schulman et al., 2015; Lillicrap et al., 2016; Silver et al., 2016; Gu et al., 2016b; Mnih et al., 2016). Deep neural network parametrization minimizes the need for manual feature and policy engineering, and allows learning end-to-end policies mapping from high-dimensional inputs, such as images, directly to actions. However, such expressive parametrization also introduces a number of practical problems. Deep reinforcement learning algorithms tend to be sensitive to hyperparameter settings, often requiring extensive hyperparameter sweeps to find good values. Poor hyperparameter settings tend to produce unstable or non-convergent learning. Deep RL algorithms also tend to exhibit high sample complexity, often to the point of being impractical to run on real physical systems. Although a number of recent techniques have sought to alleviate some of these issues (Hasselt, 2010; Mnih et al., 2015; Schulman et al., 2015; 2016), these recent advances still provide only a partial solution to the instability and sample complexity challenges.

Model-free reinforcement learning consists of on- and off-policy methods. Monte Carlo policy gradient methods (Peters & Schaal, 2006; Schulman et al., 2015) are popular on-policy methods that directly maximize the cumulative future returns with respect to the policy. While these algorithms can offer unbiased (or nearly unbiased, as discussed in Section 2.1) estimates of the gradient, they rely on Monte Carlo estimation and often suffer from high variance. To cope with high variance gradient estimates and difficult optimization landscapes, a number of techniques have been proposed, including constraining the change in the policy at each gradient step (Kakade, 2001; Peters et al., 2010) and mixing value-based back-ups to trade off bias and variance in Monte Carlo return estimates (Schulman et al., 2015). However, these methods all tend to require very large numbers of samples to deal with the high variance when estimating gradients of high-dimensional neural network policies. The crux of the problem with policy gradient methods is that they can only effectively use on-policy samples, which means that they require collecting large amounts of on-policy experiences after each parameter update to the policy. This makes them very sample intensive. Off-policy methods, such as Q-learning (Watkins & Dayan, 1992; Sutton et al., 1999; Mnih et al., 2015; Gu et al., 2016b) and off-policy actor-critic methods (Lever, 2014; Lillicrap et al., 2016), can instead use all samples, including off-policy samples, by adopting temporal difference learning with experience replay. Such methods are much more sample-efficient. However, convergence of these algorithms is in general not guaranteed with non-linear function approximators, and practical convergence and instability issues typically mean that extensive hyperparameter tuning is required to attain good results.

In order to make deep reinforcement learning practical as a tool for tackling real-world tasks, we must develop methods that are both data efficient and stable. In this paper, we propose Q-Prop, a step in this direction that combines the advantages of on-policy policy gradient methods with the efficiency of off-policy learning. Unlike prior approaches for off-policy learning, which either introduce bias (Sutton et al., 1999; Silver et al., 2014) or increase variance (Precup, 2000; Levine & Koltun, 2013; Munos et al., 2016), Q-Prop can reduce the variance of gradient estimator without adding bias; unlike prior approaches for critic-based variance reduction (Schulman et al., 2016) which fit the value function on-policy, Q-Prop learns the action-value function off-policy. The core idea is to use the first-order Taylor expansion of the critic as a control variate, resulting in an analytical gradient term through the critic and a Monte Carlo policy gradient term consisting of the residuals in advantage approximations. The method helps unify policy gradient and actor-critic methods: it can be seen as using the off-policy critic to reduce variance in policy gradient or using on-policy Monte Carlo returns to correct for bias in the critic gradient. We further provide theoretical analysis of the control variate, and derive two additional variants of Q-Prop. The method can be easily incorporated into any policy gradient algorithm. We show that Q-Prop provides substantial gains in sample efficiency over trust region policy optimization (TRPO) with generalized advantage estimation (GAE) (Schulman et al., 2015; 2016), and improved stability over deep deterministic policy gradient (DDPG) (Lillicrap et al., 2016) across a repertoire of continuous control tasks.

## 2 BACKGROUND

Reinforcement learning (RL) aims to learn a policy for an agent such that it behaves optimally according to a reward function. At a time step $t$ and state $s_t$, the agent chooses an action $a_t$ according to its policy $\pi(a_t|s_t)$, the state of the agent and the environment changes to new state $s_{t+1}$ according to dynamics $p(s_{t+1}|s_t, a_t)$, the agent receives a reward $r(s_t, a_t)$, and the process continues. Let $R_t$ denote a $\gamma$-discounted cumulative return from $t$ for an infinite horizon problem, i.e $R_t = \sum_{t'=t}^{\infty} \gamma^{t'-t} r(s_{t'}, a_{t'})$. The goal of reinforcement learning is to maximize the expected return $J(\theta) = \mathbb{E}_{\pi_\theta}[R_0]$ with respect to the policy parameters $\theta$. In this section, we review several standard techniques for performing this optimization, and in the next section, we will discuss our proposed Q-Prop algorithm that combines the strengths of these approaches to achieve efficient, stable RL. *Monte Carlo policy gradient* refers to policy gradient methods that use full Monte Carlo returns, e.g. REINFORCE (Williams, 1992) and TRPO (Schulman et al., 2015), and *policy gradient with function approximation* refers to actor-critic methods (Sutton et al., 1999) which optimize the policy against a critic, e.g. deterministic policy gradient (Silver et al., 2014; Lillicrap et al., 2016).

### 2.1 MONTE CARLO POLICY GRADIENT METHODS

Monte Carlo policy gradient methods apply direct gradient-based optimization to the reinforcement learning objective. This involves directly differentiating the $J(\theta)$ objective with respect to the policy parameters $\theta$. The standard form, known as the REINFORCE algorithm (Williams, 1992), is shown below:

$$\nabla_\theta J(\theta) = \mathbb{E}_\pi[\sum_{t=0}^\infty \nabla_\theta \log \pi_\theta(a_t|s_t) \gamma^t R_t] = \mathbb{E}_\pi[\sum_{t=0}^\infty \gamma^t \nabla_\theta \log \pi_\theta(a_t|s_t)(R_t - b(s_t))], \quad (1)$$

where $b(s_t)$ is known as the baseline. For convenience of later derivations, Eq. 1 can also be written as below, where $\rho_\pi(s) = \sum_{t=0}^\infty \gamma^t p(s_t = s)$ is the unnormalized discounted state visitation frequency,

$$\nabla_\theta J(\theta) = \mathbb{E}_{s_t \sim \rho_\pi(\cdot), a_t \sim \pi(\cdot|s_t)}[\nabla_\theta \log \pi_\theta(a_t|s_t)(R_t - b(s_t))]. \quad (2)$$

Eq. 2 is an unbiased gradient of the RL objective. However, in practice, most policy gradient methods effectively use undiscounted state visitation frequencies, i.e. $\gamma = 1$ in the equal for $\rho_\pi$, and are therefore biased; in fact, making them unbiased often hurts performance (Thomas, 2014). In this paper, we mainly discuss bias due to function approximation, off-policy learning, and value back-ups.

The gradient is estimated using Monte Carlo samples in practice and has very high variance. A proper choice of baseline is necessary to reduce the variance sufficiently such that learning becomes feasible. A common choice is to estimate the value function of the state $V_\pi(s_t)$ to use as the baseline, which provides an estimate of advantage function $A_\pi(s_t, a_t)$, which is a centered action-value function $Q_\pi(s_t, a_t)$, as defined below:

$$\begin{aligned} V_\pi(s_t) &= \mathbb{E}_\pi[R_t] = \mathbb{E}_{\pi_\theta(a_t|s_t)}[Q_\pi(s_t, a_t)] \\ Q_\pi(s_t, a_t) &= r(s_t, a_t) + \gamma \mathbb{E}_\pi[R_{t+1}] = r(s_t, a_t) + \gamma \mathbb{E}_{p(s_{t+1}|s_t, a_t)}[V_\pi(s_{t+1})] \\ A_\pi(s_t, a_t) &= Q_\pi(s_t, a_t) - V_\pi(s_t). \end{aligned} \quad (3)$$

$Q_\pi(s_t, a_t)$ summarizes the performance of each action from a given state, assuming it follows $\pi$ thereafter, and $A_\pi(s_t, a_t)$ provides a measure of how each action compares to the average performance at the state $s_t$, which is given by $V_\pi(s_t)$. Using $A_\pi(s_t, a_t)$ centers the learning signal and reduces variance significantly.

Besides high variance, another problem with the policy gradient is that it requires on-policy samples. This makes policy gradient optimization very sample intensive. To achieve similar sample efficiency as off-policy methods, we can attempt to include off-policy data. Prior attempts use importance sampling to include off-policy trajectories; however, these are known to be difficult scale to high-dimensional action spaces because of rapidly degenerating importance weights (Precup, 2000).

## 2.2 Policy Gradient with Function Approximation

Policy gradient methods with function approximation (Sutton et al., 1999), or actor-critic methods, include a *policy evaluation* step, which often uses temporal difference (TD) learning to fit a critic $Q_w$ for the current policy $\pi(\theta)$, and a *policy improvement* step which greedily optimizes the policy $\pi$ against the critic estimate $Q_w$. Significant gains in sample efficiency may be achievable using off-policy TD learning for the critic, as in Q-learning and deterministic policy gradient (Sutton, 1990; Silver et al., 2014), typically by means of *experience replay* for training deep Q networks (Mnih et al., 2015; Lillicrap et al., 2016; Gu et al., 2016b).

One particularly relevant example of such a method is the deep deterministic policy gradient (DDPG) (Silver et al., 2014; Lillicrap et al., 2016). The updates for this method are given below, where $\pi_\theta(a_t|s_t) = \delta(a_t = \mu_\theta(s_t))$ is a deterministic policy, $\beta$ is arbitrary exploration distribution, and $\rho_\beta$ corresponds to sampling from a replay buffer. $Q(\cdot, \cdot)$ is the target network that slowly tracks $Q_w$ (Lillicrap et al., 2016).

$$\begin{aligned} w &= \arg\min_w \mathbb{E}_{s_t \sim \rho_\beta(\cdot), a_t \sim \beta(\cdot|s_t)}[(r(s_t, a_t) + \gamma Q(s_{t+1}, \mu_\theta(s_{t+1})) - Q_w(s_t, a_t))^2] \\ \theta &= \arg\max_\theta \mathbb{E}_{s_t \sim \rho_\beta(\cdot)}[Q_w(s_t, \mu_\theta(s_t))] \end{aligned} \quad (4)$$

When the critic and policy are parametrized with neural networks, full optimization is expensive, and instead stochastic gradient optimization is used. The gradient in the *policy improvement* phase is given below, which is generally a biased gradient of $J(\theta)$.

$$\nabla_\theta J(\theta) \approx \mathbb{E}_{s_t \sim \rho_\beta(\cdot)}[\nabla_a Q_w(s_t, a)|_{a=\mu_\theta(s_t)} \nabla_\theta \mu_\theta(s_t)] \quad (5)$$

The crucial benefits of DDPG are that it does not rely on high variance REINFORCE gradients and is trainable on off-policy data. These properties make DDPG and other analogous off-policy methods significantly more sample-efficient than policy gradient methods (Lillicrap et al., 2016; Gu et al., 2016b; Duan et al., 2016). However, the use of a biased policy gradient estimator makes analyzing its convergence and stability properties difficult.

## 3 Q-PROP

In this section, we derive the Q-Prop estimator for policy gradient. The key idea from this estimator comes from observing Equations 2 and 5 and noting that the former provides an almost unbiased (see Section 2.1), but high variance gradient, while the latter provides a deterministic, but biased gradient. By using the deterministic biased estimator as a particular form of control variate (Ross, 2006; Paisley et al., 2012) for the Monte Carlo policy gradient estimator, we can effectively use both types of gradient information to construct a new estimator that in practice exhibits improved sample efficiency through the inclusion of off-policy samples while preserving the stability of on-policy Monte Carlo policy gradient.

### 3.1 Q-PROP ESTIMATOR

To derive the Q-Prop gradient estimator, we start by using the first-order Taylor expansion of an arbitrary function $f(s_t, a_t)$, $\bar{f}(s_t, a_t) = f(s_t, \bar{a}_t) + \nabla_a f(s_t, a)|_{a=\bar{a}_t}(a_t - \bar{a}_t)$ as the control variate for the policy gradient estimator. We use $\hat{Q}(s_t, a_t) = \sum_{t'=t}^{\infty} \gamma^{t'-t} r(s_{t'}, a_{t'})$ to denote Monte Carlo return from state $s_t$ and action $a_t$, i.e. $\mathbb{E}_\pi[\hat{Q}(s_t, a_t)] = r(s_t, a_t) + \gamma \mathbb{E}_p[V_\pi(s_{t+1})]$, and $\mu_\theta(s_t) = \mathbb{E}_{\pi_\theta(a_t|s_t)}[a_t]$ to denote the expected action of a stochastic policy $\pi_\theta$. Full derivation is in Appendix A.

$$\begin{aligned}\nabla_\theta J(\theta) &= \mathbb{E}_{\rho_\pi,\pi}[\nabla_\theta \log \pi_\theta(a_t|s_t)(\hat{Q}(s_t, a_t) - \bar{f}(s_t, a_t)] + \mathbb{E}_{\rho_\pi,\pi}[\nabla_\theta \log \pi_\theta(a_t|s_t)\bar{f}(s_t, a_t)] \\ &= \mathbb{E}_{\rho_\pi,\pi}[\nabla_\theta \log \pi_\theta(a_t|s_t)(\hat{Q}(s_t, a_t) - \bar{f}(s_t, a_t)] + \mathbb{E}_{\rho_\pi}[\nabla_a f(s_t, a)|_{a=\bar{a}_t}\nabla_\theta \mu_\theta(s_t)]\end{aligned} \quad (6)$$

Eq. 6 is general for arbitrary function $f(s_t, a_t)$ that is differentiable with respect to $a_t$ at an arbitrary value of $\bar{a}_t$; however, a sensible choice is to use the critic $Q_w$ for $f$ and $\mu_\theta(s_t)$ for $\bar{a}_t$ to get,

$$\nabla_\theta J(\theta) = \mathbb{E}_{\rho_\pi,\pi}[\nabla_\theta \log \pi_\theta(a_t|s_t)(\hat{Q}(s_t, a_t) - \bar{Q}_w(s_t, a_t))] + \mathbb{E}_{\rho_\pi}[\nabla_a Q_w(s_t, a)|_{a=\mu_\theta(s_t)}\nabla_\theta \mu_\theta(s_t)]. \quad (7)$$

Finally, since in practice we estimate advantages $\hat{A}(s_t, a_t)$, we write the Q-Prop estimator in terms of advantages to complete the basic derivation,

$$\begin{aligned}\nabla_\theta J(\theta) &= \mathbb{E}_{\rho_\pi,\pi}[\nabla_\theta \log \pi_\theta(a_t|s_t)(\hat{A}(s_t, a_t) - \bar{A}_w(s_t, a_t))] + \mathbb{E}_{\rho_\pi}[\nabla_a Q_w(s_t, a)|_{a=\mu_\theta(s_t)}\nabla_\theta \mu_\theta(s_t)] \\ \bar{A}(s_t, a_t) &= \bar{Q}(s_t, a_t) - \mathbb{E}_{\pi_\theta}[\bar{Q}(s_t, a_t)] = \nabla_a Q_w(s_t, a)|_{a=\mu_\theta(s_t)}(a_t - \mu_\theta(s_t)).\end{aligned} \quad (8)$$

Eq. 8 is composed of an analytic gradient through the critic as in Eq. 5 and a residual REINFORCE gradient in Eq. 2. From the above derivation, Q-Prop is simply a Monte Carlo policy gradient estimator with a special form of control variate. The important insight comes from the fact that $Q_w$ can be trained using off-policy data as in Eq. 4. Under this setting, Q-Prop is no longer just a Monte Carlo policy gradient method, but more closely resembles an actor-critic method, where the critic can be updated off-policy but the actor is always updated on-policy with an additional REINFORCE correction term so that it remains a Monte Carlo policy gradient method regardless of the parametrization, training method, and performance of the critic. Therefore, Q-Prop can be directly combined with a number of prior techniques from both on-policy methods such as natural policy gradient (Kakade, 2001), trust-region policy optimization (TRPO) (Schulman et al., 2015) and generalized advantage estimation (GAE) (Schulman et al., 2016), and off-policy methods such as DDPG (Lillicrap et al., 2016) and Retrace($\lambda$) (Munos et al., 2016).

Intuitively, if the critic $Q_w$ approximates $Q_\pi$ well, it provides a reliable gradient, reduces the estimator variance, and improves the convergence rate. Interestingly, control variate analysis in the next section shows that this is not the only circumstance where Q-Prop helps reduce variance.

### 3.2 Control Variate Analysis and Adaptive Q-Prop

For Q-Prop to be applied reliably, it is crucial to analyze how the variance of the estimator changes before and after the application of control variate. Following the prior work on control variates (Ross, 2006; Paisley et al., 2012), we first introduce $\eta(s_t)$ to Eq. 8, a weighing variable that modulates the strength of control variate. This additional variable $\eta(s_t)$ does not introduce bias to the estimator.

$$\nabla_\theta J(\theta) = \mathbb{E}_{\rho_\pi,\pi}[\nabla_\theta \log \pi_\theta(a_t|s_t)(\hat{A}(s_t,a_t) - \eta(s_t)\bar{A}_w(s_t,a_t))] \\ + \mathbb{E}_{\rho_\pi}[\eta(s_t)\nabla_a Q_w(s_t,a)|_{a=\mu_\theta(s_t)}\nabla_\theta \mu_\theta(s_t)] \qquad (9)$$

The variance of this estimator is given below, where $m = 1...M$ indexes the dimension of $\theta$,

$$\text{Var}^* = \mathbb{E}_{\rho_\pi}\left[\sum_m \text{Var}_{a_t}(\nabla_{\theta_m} \log \pi_\theta(a_t|s_t)(\hat{A}(s_t,a_t) - \eta(s_t)\bar{A}(s_t,a_t)))\right]. \qquad (10)$$

If we choose $\eta(s_t)$ such that $\text{Var}^* < \text{Var}$, where $\text{Var} = \mathbb{E}_{\rho_\pi}[\sum_m \text{Var}_{a_t}(\nabla_{\theta_m} \log \pi_\theta(a_t|s_t)\hat{A}(s_t,a_t))]$ is the original estimator variance measure, then we have managed to reduce the variance. Directly analyzing the above variance measure is nontrivial, for the same reason that computing the optimal baseline is difficult (Weaver & Tao, 2001). In addition, it is often impractical to get multiple action samples from the same state, which prohibits using naïve Monte Carlo to estimate the expectations. Instead, we propose a surrogate variance measure, $\text{Var} = \mathbb{E}_{\rho_\pi}[\text{Var}_{a_t}(\hat{A}(s_t,a_t))]$. A similar surrogate is also used by prior work on learning state-dependent baseline (Mnih & Gregor, 2014), and the benefit is that the measure becomes more tractable,

$$\text{Var}^* = \mathbb{E}_{\rho_\pi}[\text{Var}_{a_t}(\hat{A}(s_t,a_t) - \eta(s_t)\bar{A}(s_t,a_t))] \\ = \text{Var} + \mathbb{E}_{\rho_\pi}[-2\eta(s_t)\text{Cov}_{a_t}(\hat{A}(s_t,a_t),\bar{A}(s_t,a_t)) + \eta(s_t)^2 \text{Var}_{a_t}(\bar{A}(s_t,a_t))]. \qquad (11)$$

Since $\mathbb{E}_\pi[\hat{A}(s_t,a_t)] = \mathbb{E}_\pi[\bar{A}(s_t,a_t)] = 0$, the terms can be simplified as below,

$$\text{Cov}_{a_t}(\hat{A},\bar{A}) = \mathbb{E}_\pi[\hat{A}(s_t,a_t)\bar{A}(s_t,a_t)] \\ \text{Var}_{a_t}(\bar{A}) = \mathbb{E}_\pi[\bar{A}(s_t,a_t)^2] = \nabla_a Q_w(s_t,a)|^T_{a=\mu_\theta(s_t)}\Sigma_\theta(s_t)\nabla_a Q_w(s_t,a)|_{a=\mu_\theta(s_t)}, \qquad (12)$$

where $\Sigma_\theta(s_t)$ is the covariance matrix of the stochastic policy $\pi_\theta$. The nice property of Eq. 11 is that $\text{Var}_{a_t}(\bar{A})$ is analytical and $\text{Cov}_{a_t}(\hat{A},\bar{A})$ can be estimated with single action sample. Using this estimate, we propose adaptive variants of Q-Prop that regulate the variance of the gradient estimate.

**Adaptive Q-Prop.** The optimal state-dependent factor $\eta(s_t)$ can be computed per state, according to $\eta^*(s_t) = \text{Cov}_{a_t}(\hat{A},\bar{A})/\text{Var}_{a_t}(\bar{A})$. This provides maximum reduction in variance according to Eq. 11. Substituting $\eta^*(s_t)$ into Eq. 11, we get $\text{Var}^* = \mathbb{E}_{\rho_\pi}[(1 - \rho_{corr}(\hat{A},\bar{A})^2)\text{Var}_{a_t}(\hat{A})]$, where $\rho_{corr}$ is the correlation coefficient, which achieves guaranteed variance reduction if at any state $\bar{A}$ is correlated with $\hat{A}$. We call this the fully adaptive Q-Prop method. An important conclusion from this analysis is that, in adaptive Q-Prop, the critic $Q_w$ does not necessarily need to be approximating $Q_\pi$ well to produce good results. Its Taylor expansion merely needs to be correlated with $\hat{A}$, positively or even negatively. This is in contrast with actor-critic methods, where performance is greatly dependent on the absolute accuracy of the critic's approximation.

**Conservative and Aggressive Q-Prop.** In practice, the single-sample estimate of $\text{Cov}_{a_t}(\hat{A},\bar{A})$ has high variance itself, and we propose the following two practical implementations of adaptive Q-Prop: (1) $\eta(s_t) = 1$ if $\hat{\text{Cov}}_{a_t}(\hat{A},\bar{A}) > 0$ and $\eta(s_t) = 0$ if otherwise, and (2) $\eta(s_t) = \text{sign}(\hat{\text{Cov}}_{a_t}(\hat{A},\bar{A}))$. The first implementation, which we call conservative Q-Prop, can be thought of as a more conservative version of Q-Prop, which effectively disables the control variate for some samples of the states. This is sensible as if $\hat{A}$ and $\bar{A}$ are negatively correlated, it is likely that the critic is very poor. The second variant can correspondingly be termed aggressive Q-Prop, since it makes more liberal use of the control variate.

### 3.3 Q-Prop Algorithm

Pseudo-code for the adaptive Q-Prop algorithm is provided in Algorithm 1. It is a mixture of policy gradient and actor-critic. At each iteration, it first rolls out the stochastic policy to collect on-policy

---

**Algorithm 1** Adaptive Q-Prop
1: Initialize $w$ for critic $Q_w$, $\theta$ for stochastic policy $\pi_\theta$, and replay buffer $\mathcal{R} \leftarrow \emptyset$.
2: **repeat**
3:     **for** $e = 1, \ldots, E$ **do**                                               ▷ Collect $E$ episodes of on-policy experience using $\pi_\theta$
4:         $s_{0,e} \sim p(s_0)$
5:         **for** $t = 0, \ldots, T-1$ **do**
6:             $a_{t,e} \sim \pi_\theta(\cdot | s_{t,e})$, $s_{t+1,e} \sim p(\cdot | s_{t,e}, a_{t,e})$, $r_{t,e} = r(s_{t,e}, a_{t,e})$
7:     Add batch data $\mathcal{B} = \{s_{0:T,1:E}, a_{0:T-1,1:E}, r_{0:T-1,1:E}\}$ to replay buffer $\mathcal{R}$
8:     Take $E \cdot T$ gradient steps on $Q_w$ using $\mathcal{R}$ and $\pi_\theta$
9:     Fit $V_\phi(s_t)$ using $\mathcal{B}$
10:    Compute $\hat{A}_{t,e}$ using GAE($\lambda$) and $\bar{A}_{t,e}$ using Eq. 7
11:    Set $\eta_{t,e}$ based on Section 3.2
12:    Compute and center the learning signals $l_{t,e} = \hat{A}_{t,e} - \eta_{t,e}\bar{A}_{t,e}$
13:    Compute $\nabla_\theta J(\theta) \approx \frac{1}{ET} \sum_e \sum_t \nabla_\theta \log \pi_\theta(a_{t,e}|s_{t,e}) l_{t,e} + \eta_{t,e} \nabla_a Q_w(s_{t,e}, a)|_{a=\mu_\theta(s_{t,e})} \nabla_\theta \mu_\theta(s_{t,e})$
14:    Take a gradient step on $\pi_\theta$ using $\nabla_\theta J(\theta)$, optionally with a trust-region constraint using $\mathcal{B}$
15: **until** $\pi_\theta$ converges.

---

samples, adds the batch to a replay buffer, takes a few gradient steps on the critic, computes $\hat{A}$ and $\bar{A}$, and finally applies a gradient step on the policy $\pi_\theta$. In our implementation, the critic $Q_w$ is fitted with off-policy TD learning using the same techniques as in DDPG (Lillicrap et al., 2016):

$$w = \arg\min_w \mathbb{E}_{s_t \sim \rho_\beta(\cdot), a_t \sim \beta(\cdot|s_t)}[(r(s_t, a_t) + \gamma \mathbb{E}_\pi[Q'(s_{t+1}, a_{t+1})] - Q_w(s_t, a_t))^2]. \quad (13)$$

$V_\phi$ is fitted with the same technique in (Schulman et al., 2016). Generalized advantage estimation (GAE) (Schulman et al., 2016) is used to estimate $\hat{A}$. The policy update can be done by any method that utilizes the first-order gradient and possibly the on-policy batch data, which includes trust region policy optimization (TRPO) (Schulman et al., 2015). Importantly, this is just one possible implementation of Q-Prop, and in Appendix C we show a more general form that can interpolate between pure policy gradient and off-policy actor-critic.

### 3.4 LIMITATIONS

A limitation with Q-Prop is that if data collection is very fast, e.g. using fast simulators, the compute time per episode is bound by the critic training at each iteration, and similar to that of DDPG and usually much more than that of TRPO. However, in applications where data collection speed is the bottleneck, there is sufficient time between policy updates to fit $Q_w$ well, which can be done asynchronously from the data collection, and the compute time of Q-Prop will be about the same as that of TRPO.

Another limitation is the robustness to bad critics. We empirically show that our conservative Q-Prop is more robust than standard Q-Prop and much more robust than pure off-policy actor-critic methods such as DDPG; however, estimating when an off-policy critic is reliable or not is still a fundamental problem that shall be further investigated. We can also alleviate this limitation by adopting more stable off-policy critic learning techniques such as Retrace($\lambda$) (Munos et al., 2016).

## 4 RELATED WORK

Variance reduction in policy gradient methods is a long-standing problem with a large body of prior work (Weaver & Tao, 2001; Greensmith et al., 2004; Schulman et al., 2016). However, exploration of action-dependent control variates is relatively recent, with most work focusing instead on simpler baselining techniques (Ross, 2006). A subtle exception is compatible feature approximation (Sutton et al., 1999) which can be viewed as a control variate as explained in Appendix B. Another exception is doubly robust estimator in contextual bandits (Dudík et al., 2011), which uses a different control variate whose bias cannot be tractably corrected. Control variates were explored recently not in RL but for approximate inference in stochastic models (Paisley et al., 2012), and the closest related work in that domain is the MuProp algorithm (Gu et al., 2016a) which uses a mean-field network as a surrogate for backpropagating a deterministic gradient through stochastic discrete variables. MuProp is not directly applicable to model-free RL because the dynamics are unknown; however, it can be if the dynamics are learned as in model-based RL (Atkeson & Santamaria, 1997; Deisenroth & Rasmussen, 2011). This model-based Q-Prop is itself an interesting direction of research as it effectively corrects bias in model-based learning.

Part of the benefit of Q-Prop is the ability to use off-policy data to improve on-policy policy gradient methods. Prior methods that combine off-policy data with policy gradients either introduce bias (Sutton et al., 1999; Silver et al., 2014) or use importance weighting, which is known to result in degenerate importance weights in high dimensions, resulting in very high variance (Precup, 2000; Levine & Koltun, 2013). Q-Prop provides a new approach for using off-policy data to reduce variance without introducing further bias.

Lastly, since Q-Prop uses both on-policy policy updates and off-policy critic learning, it can take advantage of prior work along both lines of research. We chose to implement Q-Prop on top of TRPO-GAE primarily for the purpose of enabling a fair comparison in the experiments, but combining Q-Prop with other on-policy update schemes and off-policy critic training methods is an interesting direction for future work. For example, Q-Prop can also be used with other on-policy policy gradient methods such as A3C (Mnih et al., 2016) and off-policy advantage estimation methods such as Retrace($\lambda$) (Munos et al., 2016), GTD2 (Sutton et al., 2009), emphatic TD (Sutton et al., 2015), and WIS-LSTD (Mahmood et al., 2014).

## 5 EXPERIMENTS

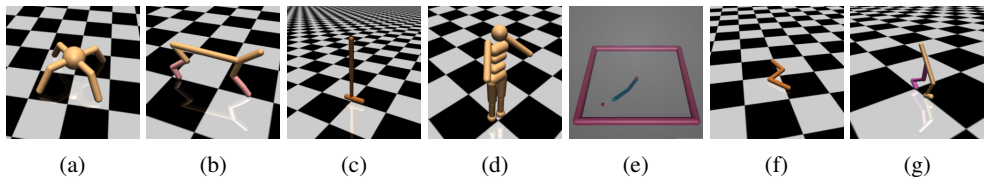

(a)      (b)      (c)      (d)      (e)      (f)      (g)

Figure 1: Illustrations of OpenAI Gym MuJoCo domains (Brockman et al., 2016; Duan et al., 2016): (a) Ant, (b) HalfCheetah, (c) Hopper, (d) Humanoid, (e) Reacher, (f) Swimmer, (g) Walker.

We evaluated Q-Prop and its variants on continuous control environments from the OpenAI Gym benchmark (Brockman et al., 2016) using the MuJoCo physics simulator (Todorov et al., 2012) as shown in Figure 1. Algorithms are identified by acronyms, followed by a number indicating batch size, except for DDPG, which is a prior online actor-critic algorithm (Lillicrap et al., 2016). "c-" and "v-" denote conservative and aggressive Q-Prop variants as described in Section 3.2. "TR-" denotes trust-region policy optimization (Schulman et al., 2015), while "V-" denotes vanilla policy gradient. For example, "TR-c-Q-Prop-5000" means convervative Q-Prop with the trust-region policy update, and a batch size of 5000. "VPG" and "TRPO" are vanilla policy gradient and trust-region policy optimization respectively (Schulman et al., 2016; Duan et al., 2016). Unless otherwise stated, all policy gradient methods are implemented with GAE($\lambda = 0.97$) (Schulman et al., 2016). Note that TRPO-GAE is currently the state-of-the-art method on most of the OpenAI Gym benchmark tasks, though our experiments show that a well-tuned DDPG implementation sometimes achieves better results. Our algorithm implementations are built on top of the rllab TRPO and DDPG codes from Duan et al. (2016) and available at https://github.com/shaneshixiang/rllabplusplus. Policy and value function architectures and other training details including hyperparameter values are provided in Appendix D.

### 5.1 ADAPTIVE Q-PROP

First, it is useful to identify how reliable each variant of Q-Prop is. In this section, we analyze standard Q-Prop and two adaptive variants, c-Q-Prop and a-Q-Prop, and demonstrate the stability of the method across different batch sizes. Figure 2a shows a comparison of Q-Prop variants with trust-region updates on the HalfCheetah-v1 domain, along with the best performing TRPO hyperparameters. The results are consistent with theory: conservative Q-Prop achieves much more stable performance than the standard and aggressive variants, and all Q-Prop variants significantly outperform TRPO in terms of sample efficiency, e.g. conservative Q-Prop reaches average reward of 4000 using about 10 times less samples than TRPO.

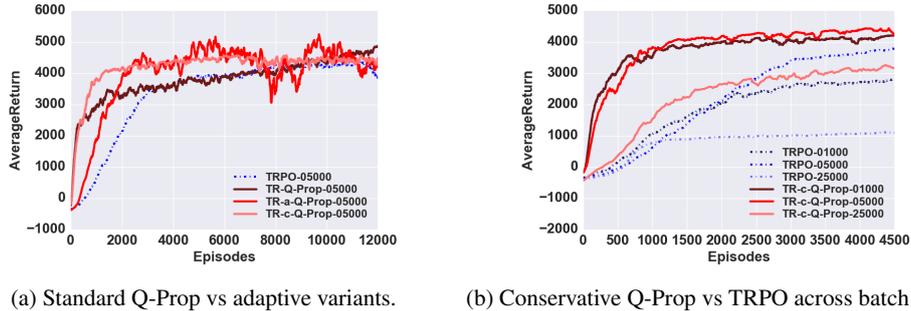

(a) Standard Q-Prop vs adaptive variants.

(b) Conservative Q-Prop vs TRPO across batch sizes.

Figure 2: Average return over episodes in HalfCheetah-v1 during learning, exploring adaptive Q-Prop methods and different batch sizes. All variants of Q-Prop substantially outperform TRPO in terms of sample efficiency. TR-c-QP, conservative Q-Prop with trust-region update performs most stably across different batch sizes.

Figure 2b shows the performance of conservative Q-Prop against TRPO across different batch sizes. Due to high variance in gradient estimates, TRPO typically requires very large batch sizes, e.g. 25000 steps or 25 episodes per update, to perform well. We show that our Q-Prop methods can learn even with just 1 episode per update, and achieves better sample efficiency with small batch sizes. This shows that Q-Prop significantly reduces the variance compared to the prior methods.

As we discussed in Section 1, stability is a significant challenge with state-of-the-art deep RL methods, and is very important for being able to reliably use deep RL for real world tasks. In the rest of the experiments, we will use conservative Q-Prop as the main Q-Prop implementation.

## 5.2 EVALUATION ACROSS ALGORITHMS

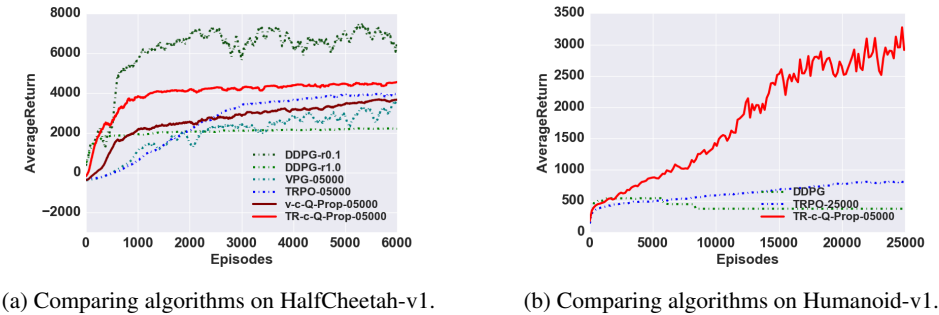

(a) Comparing algorithms on HalfCheetah-v1.

(b) Comparing algorithms on Humanoid-v1.

Figure 3: Average return over episodes in HalfCheetah-v1 and Humanoid-v1 during learning, comparing Q-Prop against other model-free algorithms. Q-Prop with vanilla policy gradient outperforms TRPO on HalfCheetah. Q-Prop significantly outperforms TRPO in convergence time on Humanoid.

In this section, we evaluate two versions of conservative Q-Prop, v-c-Q-Prop using vanilla policy gradient and TR-c-Q-Prop using trust-region updates, against other model-free algorithms on the HalfCheetah-v1 domain. Figure 3a shows that c-Q-Prop methods significantly outperform the best TRPO and VPG methods. Even Q-Prop with vanilla policy gradient is comparable to TRPO, confirming the significant benefits from variance reduction. DDPG on the other hand exhibits inconsistent performances. With proper reward scaling, i.e. "DDPG-r0.1", it outperforms other methods as well as the DDPG results reported in prior work (Duan et al., 2016; Amos et al., 2016). This illustrates the sensitivity of DDPG to hyperparameter settings, while Q-Prop exhibits more stable, monotonic learning behaviors when compared to DDPG. In the next section we show this improved stability allows Q-Prop to outperform DDPG in more complex domains.

### 5.3 EVALUATION ACROSS DOMAINS

Lastly, we evaluate Q-Prop against TRPO and DDPG across multiple domains. While the gym environments are biased toward locomotion, we expect we can achieve similar performance on manipulation tasks such as those in Lillicrap et al. (2016). Table 1 summarizes the results, including the best attained average rewards and the steps to convergence. Q-Prop consistently outperform TRPO in terms of sample complexity and sometimes achieves higher rewards than DDPG in more complex domains. A particularly notable case is shown in Figure 3b, where Q-Prop substantially improves sample efficiency over TRPO on Humanoid-v1 domain, while DDPG cannot find a good solution.

The better performance on the more complex domains highlights the importance of stable deep RL algorithms: while costly hyperparameter sweeps may allow even less stable algorithms to perform well on simpler problems, more complex tasks might have such narrow regions of stable hyperparameters that discovering them becomes impractical.

|  |  | TR-c-Q-Prop |  | TRPO |  | DDPG |  |
| --- | --- | --- | --- | --- | --- | --- | --- |
| Domain | Threshold | MaxReturn. | Episodes | MaxReturn | Epsisodes | MaxReturn | Episodes |
| Ant | 3500 | 3534 | **4975** | **4239** | 13825 | 957 | N/A |
| HalfCheetah | 4700 | 4811 | 20785 | 4734 | 26370 | **7490** | **600** |
| Hopper | 2000 | **2957** | 5945 | 2486 | 5715 | 2604 | **965** |
| Humanoid | 2500 | **>3492** | **14750** | 918 | >30000 | 552 | N/A |
| Reacher | -7 | **-6.0** | 2060 | -6.7 | 2840 | -6.6 | **1800** |
| Swimmer | 90 | 103 | 2045 | 110 | 3025 | **150** | **500** |
| Walker | 3000 | **4030** | 3685 | 3567 | 18875 | 3626 | **2125** |

Table 1: Q-Prop, TRPO and DDPG results showing the max average rewards attained in the first 30k episodes and the episodes to cross specific reward thresholds. Q-Prop often learns more sample efficiently than TRPO and can solve difficult domains such as Humanoid better than DDPG.

## 6 DISCUSSION AND CONCLUSION

We presented Q-Prop, a policy gradient algorithm that combines reliable, consistent, and potentially unbiased on-policy gradient estimation with a sample-efficient off-policy critic that acts as a control variate. The method provides a large improvement in sample efficiency compared to state-of-the-art policy gradient methods such as TRPO, while outperforming state-of-the-art actor-critic methods on more challenging tasks such as humanoid locomotion. We hope that techniques like these, which combine on-policy Monte Carlo gradient estimation with sample-efficient variance reduction through off-policy critics, will eventually lead to deep reinforcement learning algorithms that are more stable and efficient, and therefore better suited for application to complex real-world learning tasks.

#### ACKNOWLEDGMENTS

We thank Rocky Duan for sharing and answering questions about rllab code, and Yutian Chen and Laurent Dinh for discussion on control variates. SG and RT were funded by NSERC, Google, and EPSRC grants EP/L000776/1 and EP/M026957/1. ZG was funded by EPSRC grant EP/J012300/1 and the Alan Turing Institute (EP/N510129/1).

## A  Q-PROP ESTIMATOR DERIVATION

The full derivation of the Q-Prop estimator is shown in Eq. 14. We make use of the following property that is commonly used in baseline derivations:

$$\mathbb{E}_{p_\theta(x)}[\nabla_\theta \log p_\theta(x)] = \int_x \nabla_\theta p_\theta(x) = \nabla_\theta \int_x p(x) = 0$$

This holds true when $f(s_t, a_t)$ is an arbitrary function differentiable with respect to $a_t$ and $\bar{f}$ is its first-order Taylor expansion around $a_t = \bar{a}_t$, i.e. $\bar{f}(s_t, a_t) = f(s_t, \bar{a}_t) + \nabla_a f(s_t, a)|_{a=\bar{a}_t}(a_t - \bar{a}_t)$. Here, $\mu_\theta(s_t) = \mathbb{E}_\pi[a_t]$ is the mean of stochastic policy $\pi_\theta$. The derivation appears below:

$$\begin{aligned}
\nabla_\theta J(\theta) &= \mathbb{E}_{\rho_\pi,\pi}[\nabla_\theta \log \pi_\theta(a_t|s_t)(\hat{Q}(s_t,a_t) - \bar{f}(s_t,a_t)] + \mathbb{E}_{\rho_\pi,\pi}[\nabla_\theta \log \pi_\theta(a_t|s_t)\bar{f}(s_t,a_t)] \\
g(\theta) &= \mathbb{E}_{\rho_\pi,\pi}[\nabla_\theta \log \pi_\theta(a_t|s_t)\bar{f}(s_t,a_t)] \\
&= \mathbb{E}_{\rho_\pi,\pi}[\nabla_\theta \log \pi_\theta(a_t|s_t)(f(s_t,\bar{a}_t) + \nabla_a f(s_t,a)|_{a=\bar{a}_t}(a_t - \bar{a}_t))] \\
&= \mathbb{E}_{\rho_\pi,\pi}[\nabla_\theta \log \pi_\theta(a_t|s_t)\nabla_a f(s_t,a)|_{a=\bar{a}_t} a_t] \\
&= \mathbb{E}_{\rho_\pi}\left[\int_{a_t} \nabla_\theta \pi_\theta(a_t|s_t) \nabla_a f(s_t,a)|_{a=\bar{a}_t} a_t\right] \\
&= \mathbb{E}_{\rho_\pi}\left[\nabla_a f(s_t,a)|_{a=\bar{a}_t} \int_{a_t} \nabla_\theta \pi_\theta(a_t|s_t) a_t\right] \\
&= \mathbb{E}_{\rho_\pi}[\nabla_a f(s_t,a)|_{a=\bar{a}_t} \nabla_\theta \mathbb{E}_\pi[a_t]] \\
&= \mathbb{E}_{\rho_\pi}[\nabla_a f(s_t,a)|_{a=\bar{a}_t} \nabla_\theta \mu_\theta(s_t)] \\
\nabla_\theta J(\theta) &= \mathbb{E}_{\rho_\pi,\pi}[\nabla_\theta \log \pi_\theta(a_t|s_t)(\hat{Q}(s_t,a_t) - \bar{f}(s_t,a_t)] + g(\theta) \\
&= \mathbb{E}_{\rho_\pi,\pi}[\nabla_\theta \log \pi_\theta(a_t|s_t)(\hat{Q}(s_t,a_t) - \bar{f}(s_t,a_t)] + \mathbb{E}_{\rho_\pi}[\nabla_a f(s_t,a)|_{a=\bar{a}_t} \nabla_\theta \mu_\theta(s_t)]
\end{aligned} \quad (14)$$

## B  Connection Between Q-Prop and Compatible Feature Approximation

In this section we show that actor-critic with compatible feature approximation is a form of control variate. A critic $Q_w$ is *compatible* (Sutton et al., 1999) if it satisfies (1) $Q_w(s_t, a_t) = w^T \nabla_\theta \log \pi_\theta(a_t|s_t)$, i.e. $\nabla_w Q_w(s_t, a_t) = \nabla_\theta \log \pi_\theta(a_t|s_t)$, and (2) $w$ is fit with objective $w = \arg\min_w L(w) = \arg\min_w \mathbb{E}_{\rho_\pi,\pi}[(\hat{Q}(s_t,a_t) - Q_w(s_t,a_t))^2]$, that is fitting $Q_w$ on on-policy Monte Carlo returns. Condition (2) implies the following identity,

$$\nabla_w L = 2\mathbb{E}_{\rho_\pi,\pi}[\nabla_\theta \log \pi_\theta(a_t|s_t)(\hat{Q}(s_t,a_t) - Q_w(s_t,a_t))] = 0. \quad (15)$$

In compatible feature approximation, it directly uses $Q_w$ as control variate, rather than its Taylor expansion $\bar{Q}_w$ as in Q-Prop. Using Eq. 15, the Monte Carlo policy gradient is,

$$\begin{aligned}
\nabla_\theta J(\theta) &= \mathbb{E}_{\rho_\pi,\pi}[\nabla_\theta \log \pi_\theta(a_t|s_t) Q_w(s_t,a_t)] \\
&= \mathbb{E}_{\rho_\pi,\pi}[(\nabla_\theta \log \pi_\theta(a_t|s_t) \nabla_\theta \log \pi_\theta(a_t|s_t)^T) w] \\
&= \mathbb{E}_{\rho_\pi}[I(\theta; s_t) w],
\end{aligned} \quad (16)$$

where $I(\theta; s_t) = \mathbb{E}_{\pi_\theta}[\nabla_\theta \log \pi_\theta(a_t|s_t) \nabla_\theta \log \pi_\theta(a_t|s_t)^T]$ is Fisher's information matrix. Thus, variance reduction depends on ability to compute or estimate $I(\theta; s_t)$ and $w$ effectively.

## C  Unifying Policy Gradient and Actor-Critic

Q-Prop closely ties together policy gradient and actor-critic algorithms. To analyze this point, we write a generalization of Eq. 9 below, introducing two additional variables $\alpha, \rho_{CR}$:

$$\begin{aligned}
\nabla_\theta J(\theta) \propto &\alpha \mathbb{E}_{\rho_\pi,\pi}[\nabla_\theta \log \pi_\theta(a_t|s_t)(\hat{A}(s_t,a_t) - \eta \bar{A}_w(s_t,a_t)] \\
&+ \eta \mathbb{E}_{\rho_{CR}}[\nabla_a Q_w(s_t,a)|_{a=\mu_\theta(s_t)} \nabla_\theta \mu_\theta(s_t)]
\end{aligned} \quad (17)$$

Eq. 17 enables more analysis where bias generally is introduced only when $\alpha \neq 1$ or $\rho_{CR} \neq \rho_\pi$. Importantly, Eq. 17 covers both policy gradient and deterministic actor-critic algorithm as its special cases. Standard policy gradient is recovered by $\eta = 0$, and deterministic actor-critic is recovered by $\alpha = 0$ and $\rho_{CR} = \rho_\beta$. This allows heuristic or automatic methods for dynamically changing these variables through the learning process for optimizing different metrics, e.g. sample efficiency, convergence speed, stability.

Table 2 summarizes the various edge cases of Eq. 17. For example, since we derive our method from a control variates standpoint, $Q_w$ can be any function and the gradient remains almost unbiased (see

| Parameter | Implementation options | Introduce bias? |
|---|---|---|
| $Q_w$ | off-policy TD; on-policy TD($\lambda$); model-based; etc. | No |
| $V_\phi$ | on-policy Monte Carlo fitting; $\mathbb{E}_{\pi_\theta}[Q_w(s_t, a_t)]$; etc | No |
| $\lambda$ | $0 \leq \lambda \leq 1$ | Yes, except $\lambda = 1$ |
| $\alpha$ | $\alpha \geq 0$ | Yes, except $\alpha = 1$ |
| $\eta$ | any $\eta$ | No |
| $\rho_{CR}$ | $\rho$ of any policy | Yes, except $\rho_{CR} = \rho_\pi$ |

Table 2: Implementation options and edge cases of the generalized Q-Prop estimator in Eq. 17.

Section 2.1). A natural choice is to use off-policy temporal difference learning to learn the critic $Q_w$ corresponding to policy $\pi$. This enables effectively utilizing off-policy samples without introducing further bias. An interesting alternative to this is to utilize model-based roll-outs to estimate the critic, which resembles MuProp in stochastic neural networks (Gu et al., 2016a). Unlike prior work on using fitted dynamics model to accelerate model-free learning (Gu et al., 2016b), this approach does not introduce bias to the gradient of the original objective.

## D  EXPERIMENT DETAILS

**Policy and value function architectures.** The network architectures are largely based on the benchmark paper by Duan et al. (2016). For policy gradient methods, the stochastic policy $\pi_\theta(a_t|s_t) = \mathcal{N}(\mu_\theta(s_t), \Sigma_\theta)$ is a local Gaussian policy with a local state-dependent mean and a global covariance matrix. $\mu_\theta(s_t)$ is a neural network with 3 hidden layers of sizes 100-50-25 and tanh nonlinearities at the first 2 layers, and $\Sigma_\theta$ is diagonal. For DDPG, the policy is deterministic and has the same architecture as $\mu_\theta$ except that it has an additional tanh layer at the output. $V_\phi(s_t)$ for baselines and GAE is fit with the same technique by Schulman et al. (2016), a variant of linear regression on Monte Carlo returns with soft-update constraint. For Q-Prop and DDPG, $Q_w(s, a)$ is parametrized with a neural network with 2 hidden layers of size 100 and ReLU nonlinearity, where $a$ is included after the first hidden layer.

**Training details.** This section describes parameters of the training algorithms and their hyperparameter search values in {}. The optimal performing hyperparameter results are reported. Policy gradient methods (VPG, TRPO, Q-Prop) used batch sizes of {1000, 5000, 25000} time steps, step sizes of {0.1, 0.01, 0.001} for the trust-region method, and base learning rates of {0.001, 0.0001} with Adam (Kingma & Ba, 2014) for vanilla policy gradient methods. For Q-Prop and DDPG, $Q_w$ is learned with the same technique as in DDPG (Lillicrap et al., 2016), using soft target networks with $\tau = 0.999$, a replay buffer of size $10^6$ steps, a mini-batch size of 64, and a base learning rate of {0.001, 0.0001} with Adam (Kingma & Ba, 2014). For Q-Prop we also tuned the relative ratio of gradient steps on the critic $Q_w$ against the number of steps on the policy, in the range {0.1, 0.5, 1.0}, where 0.1 corresponds to 100 critic updates for every policy update if the batch size is 1000. For DDPG, we swept the reward scaling using {0.01, 0.1, 1.0} as it is sensitive to this parameter.